# Cuspidal and Noncuspidal Robot Manipulators

Philippe Wenger

Institut de Recherche en Communications et Cybernétique de Nantes

1, rue de la Noë, 44321 Nantes, France

Email: Philippe.Wenger@irccyn.ec-nantes.fr

**Abstract**

This article synthezises the most important results on the kinematics of cuspidal manipulators i.e. nonredundant manipulators that can change posture without meeting a singularity. The characteristic surfaces, the uniqueness domains and the regions of feasible paths in the workspace are defined. Then, several sufficient geometric conditions for a manipulator to be noncuspidal are enumerated and a general necessary and sufficient condition for a manipulator to be cuspidal is provided. An explicit DH-parameter-based condition for an orthogonal manipulator to be cuspidal is derived. The full classification of 3R orthogonal manipulators is provided and all types of cuspidal and noncuspidal orthogonal manipulators are enumerated. Finally, some facts about cuspidal and noncuspidal 6R manipulators are reported.

## 1   Introduction

A cuspidal manipulator is a nonredundant manipulator that can change its posture (a posture is associated with an inverse kinematic solution) without meeting a singularity. Today, most industrial 6R manipulators are of the PUMA type, which is noncuspidal. Indeed, a Puma robot cannot avoid the fully extended arm configuration when moving from the "elbow up" to the "elbow down" posture. Modification in the link arrangement is very likely to result in a cuspidal manipulator. In 1998, ABB-Robotics launched the IRB 6400C, a new manipulator specially designed for the car industry to minimize the swept volume. The only difference from the Puma was the permutation of the first two link axes, resulting in a manipulator with all its joint axes orthogonal. Commercialization of the IRB 6400C was finally stopped one year later. Informal interviews with robot customers at that time revealed difficulties in planning offline trajectories using Robotic-CAD systems for this robot. In fact it turns out that the IRB 6400C robot is cuspidal. We will come back to this robot in section 5.

It has long been believed that any manipulator always encounters a singularity during a change of posture[1]. The nonsingular change of posture was first pointed out in 1988 in two separate works. Parenti-Castelli and Innocenti exhibited a nonsingular posture changing trajectory for two different 6R cuspidal manipulators using numerical experiments[2] while Burdick provided several examples of cuspidal 3R manipulators and some general results about which manipulators should be cuspidal[3]. Maybe because of the scepticism of the research community at that time, this feature has been ignored for several years and no further research work was provided before 1992, when the nonsingular posture-changing ability was confirmed and more formally analyzed[4]. As few authors have investigated this phenomenon since then, it took a long time before the research community recognized the nonsingular posture-changing ability. The problem of planning non-singular changing posture trajectories for general 3R manipulators was addressed by Tsai and Kholi in 1993[5]. At the very end of his work, Smith suggested that a non-singular posture changing trajectory should encircle a cusp point in the workspace[6]. Burdick provided a list of conditions on the DH-parameters for a manipulator to be noncuspidal[7]. This list includes simplifying geometric conditions such as parallel and intersecting joint axes. Later, Wenger provided other conditions that are not intuitive[8]. A general, necessary and sufficient condition for a 3-DOF manipulator to be cuspidal was first established by El Omri and Wenger[9] in 1995, namely, the existence of at least one point in the workspace where the inverse kinematics admits three equal solutions. The word "cuspidal manipulator" was defined in accordance to this condition because a point with three equal inverse kinematic solution forms a cusp in a cross section of the workspace[7,10]. The different possible posture-changing motions for 3-DOF manipulators were analyzed by Wenger and El Omri[11]. The categorization of all generic 3R manipulators was established using homotopy classes, which made it possible to show that the space of 3R manipulators is mostly composed of cuspidal ones[12]. A procedure to take into account the cuspidality property in the design process of new manipulators was provided[13]. More recently, Corvez and Rouillier attempted the classification of 3R manipulators with orthogonal joints[14]. Five surfaces were found to divide the manipulator parameter space into cells with constant number of cusp points. The equations of these surfaces were derived as polynomials in the DH-parameters using Groebner Bases. A physical interpretation of this theoretical work was conducted by Baili et al[15] who pointed out the existence of extraneous surface equations and took into account additional features in the classification like genericity[16] and the number of aspects. The complete classification of orthogonal 3R manipulators was established for the first time in 2004 on the basis of the number of cusps and nodes in the workspace cross section[17, 18]. A general formalism for the kinematic analysis of cuspidal manipulators was provided and the maximal sets of feasible paths in the workspace were defined[19].

The purpose of this work is to synthesize the most important results on the kinematics of cuspidal manipulators i.e. nonredundant manipulators that can change posture without meeting a singularity.

The rest of this article is organized as follows. Section 2 introduces an illustrative cuspidal manipulator and recalls some facts about singularities and aspects. Section 3 defines the characteristic surfaces, the uniqueness domains and the regions of feasible paths in the workspace. Section 4 is devoted to the classification and enumeration of cuspidal and noncuspidal manipulators. The last section addresses the case of 6R manipulators.

## 2    Preliminaries

*2.1    Illustrative Manipulator*

A typical 3R cuspidal manipulator is used as illustrative example in this section. A 3R cuspidal manipulator should not have parallel or intersecting joint axes to be cuspidal[7, 8]. The geometric parameters of this manipulator, known as DH-parameters, are taken as $\alpha_1=-\pi/2$, $\alpha_2=\pi/2$, $a_1=1$, $a_2=2$, $a_3=1.5$, $d_1=0$, $d_2=1$, $d_3=0$. This manipulator with mutually orthogonal joint axes (henceforth referred to as an *orthogonal manipulator*) has a rather simple geometry and is a good representative example[11, 13]. The three joint variables are referred to as $\theta_1$, $\theta_2$ and $\theta_3$, respectively. Fig. 1 shows the kinematic architecture of the manipulator in its zero configuration, i.e. $\theta_1 = \theta_2 = \theta_3 = 0$. This manipulator can be regarded as the regional structure of a 6R robot with a spherical wrist. The position of the end-tip is defined by the three Cartesian coordinates *px*, *py* and *pz* of the operation point P with respect to a reference frame (O, **X**, **Y**, **Z**) attached to the manipulator base (Fig. 1).

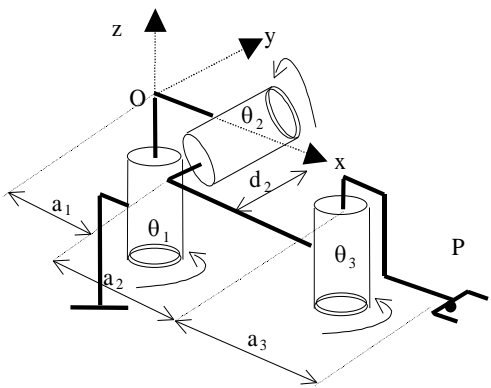
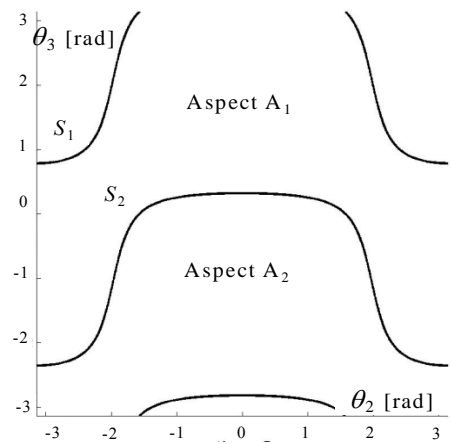

**Fig. 1. A cuspidal 3R manipulator.**              **Fig. 2. Aspects of the cuspidal manipulator.**

*2.2    Singularities and aspects*

The singularities of a manipulator play an important role in its global kinematic properties[3-7, 1-22]. The singularities of a 3R manipulator can be determined using a recursive appoach[23] or with det(**J**), the determinant of the Jacobian matrix $\delta(px,py,pz)/\delta(\theta_1,\theta_2,\theta_3)$[7,8,12,15]. This determinant can be derived automatically with symbolic softwares such as SYMORO[24]. For the illustrative manipulator, det(**J**) takes the following factored form,

$$\det(\mathbf{J}) = (a_2 + c_3 a_3)(c_2(s_3 a_2 - c_3 d_2) + s_3 a_1) \tag{1}$$

where $c_i=\cos(\theta_i)$ and $s_i=\sin(\theta_i)$. A singularity occurs when det(**J**)=0. Since the singularities are independent of $\theta_1$, the contour plot of det(**J**)=0 can be displayed in $-\pi \leq \theta_2 < \pi, -\pi \leq \theta_3 < \pi$. For the illustrative manipulator $a_2>a_3$ and the first factor of det(**J**) cannot vanish (examples with $a_2<a_3$ will be shown in Fig. 11 in section 4.3). Fig. 2 shows that the singularities form two closed surfaces $S_1$ and $S_2$ in the joint space. If the manipulator has no joint limits, $S_1$ and $S_2$ divide the joint space into two singularity-free open, connected sets $A_1$ and $A_2$ called c-sheets[3] or *aspects*[1]. We use the term 'aspect' because it is also used for manipulators with limited joints whereas c-sheets were defined for manipulators with unlimited joints only.

## 2.3  Singularities and workspace

The workspace of general 3R manipulators has been widely studied since the seventies[1,3-8, 17-22, 25-36]. The determination of the workspace boundaries, the size and shape of the workspace, the existence of holes and voids, accessibility inside the workspace (i.e. the number of inverse kinematic solutions) are some of the main features that have been explored. The singularities can be displayed in Cartesian space where they define boundaries. Thanks to their symmetry about the first joint axis, a representation in a half cross-section of the workspace is sufficient (Fig. 3). As in the joint space, the singularities also form two disjoint curves in the workspace. These two curves define the internal boundary $WS_1$ and the external boundary $WS_2$, respectively. If **f** denotes the kinematic map, then $WS_1=\mathbf{f}(S_1)$ and $WS_2=\mathbf{f}(S_2)$. The separating and sorting of these boundary curves have been recently studied in detail[36]. The internal boundary $WS_1$ is composed of four adjacent arcs $BS_1$, $BS_2$, $BS_3$ and $BS_4$ connected by four cusp points. It divides the workspace into one region with two inverse kinematic solutions (the outer region) and one region with four inverse kinematic solutions (the inner region, Fig. 3). Each point on the internal boundary has three distinct inverse kinematic solutions, one of which is a double solution. At each cusp point, there are only two distinct inverse kinematic solutions, one of them being a triple solution[4]. The external boundary surface is composed of two adjacent arcs that meet on axis **Z**. There is only one inverse kinematic solution on the external boundary, which is a double solution. There are an infinite number of inverse kinematic solutions at the two connecting points on axis **Z** (because $\theta_1$ can take on any value without altering the position of the end-tip).

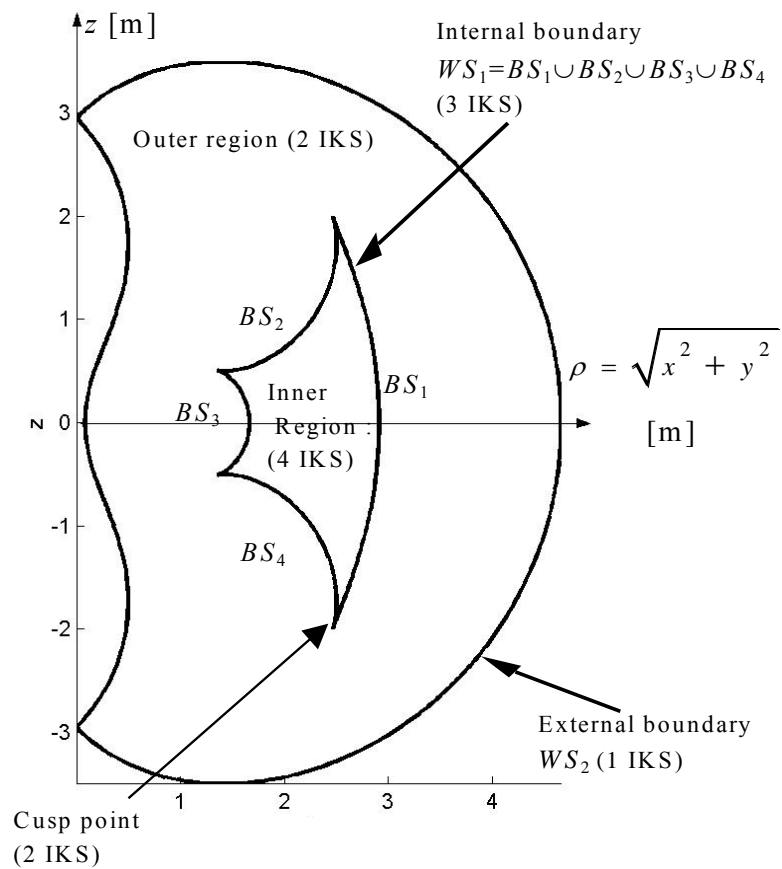

**Fig. 3. Singularity locus in workspace. Number of distinct inverse kinematic solutions (IKS) in each region and on each boundary is indicated.**

## 2.4 *A nonsingular posture changing trajectory*

For the illustrative manipulator, solving the inverse kinematics at *px*=2.5, *py*=0, *pz*=0.5 yields four solutions given (in radians) by $\mathbf{q}^{(1)}$=[-1.8 -2.8 1.9]$^t$, $\mathbf{q}^{(2)}$=[-0.9 -0.7 2.5]$^t$, $\mathbf{q}^{(3)}$=[-2.9 -3 -0.2]$^t$ and $\mathbf{q}^{(4)}$=[0.2 –0.3 –1.9]$^t$. It is apparent from fig. 4 that $\mathbf{q}^{(2)}$ and $\mathbf{q}^{(3)}$ (resp. $\mathbf{q}^{(1)}$ and $\mathbf{q}^{(4)}$ ) lie in the same aspect $A_1$ (resp. $A_2$), which means that these two solutions are not separated by a singularity. It is then possible to link $\mathbf{q}^{(2)}$ and $\mathbf{q}^{(3)}$ by a nonsingular straight line trajectory. When projected in the workspace cross section, this trajectory traces a loop path that encompasses a cusp point (Fig. 4). In fact it has been shown that a nonsingular posture-changing trajectory always encompasses a cusp point in the workspace[4-6, 9, 11].

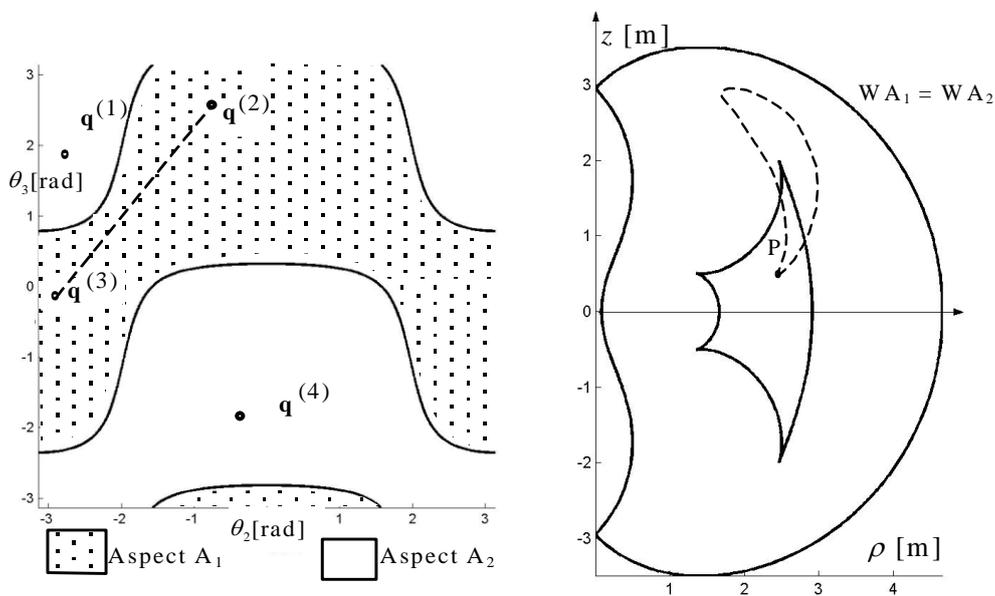

**Fig. 4. A point with two inverse kinematic solutions in each aspect. A nonsingular posture changing trajectory and the resulting path in workspace section are displayed.**

On the other hand, there is only one inverse kinematic solution per aspect for any point in the outer region. The two aspects $A_1$ and $A_2$ map onto the same set $WA_1$=$WA_2$ in the workspace, with the same internal and external boundaries as in Fig. 4. In $WA_1$ or in $WA_2$, there are only two solutions in the inner region and one solution in the outer region.

# 3  Formalism for the kinematic analysis of cuspidal manipulators

In this section, the notion of characteristic surfaces and uniqueness domains is introduced. Not every motion is feasible in the workspace of a cuspidal manipulator, even without joint limits. It is shown that the regions of feasible motions in the workspace are defined as the image of the uniqueness domains through the kinematic map. This holds for any nonredundant manipulator with or without joint limits.

## *3.1  Characteristic surfaces*

Since the singular surfaces in the joint space do not separate all the inverse kinematic solutions, new separating surfaces should exist. The set obtained by calculating the nonsingular inverse kinematic solutions for all points on an internal boundary forms a set of nonsingular surfaces in each aspect. We call these surfaces the *characteristic surfaces*. A set of characteristic surfaces is associated with one aspect and it was proved that they separate the inverse kinematic solutions in each aspect[17]. A general definition of the characteristic surfaces can be set as follows, which stands for any nonredundant manipulator with or without joint limits. Let $A_i*$ be the boundary of aspect $A_i$. The characteristic surfaces $\{CS_i\}$ associated with $A_i$ are :

$$\{CS_i\} = f^{-1}(f(A_i*)) \cap A_i \qquad (2)$$

where $f(A_i*)$ is the image of $A_i*$ under the forward kinematic map and $f^{-1}(f(A_i*)) = \{\mathbf{q} / f(\mathbf{q}) \in f(A_i*)\}$. Note that since an aspect is defined as an open set, $A_i$ does not contain its boundary i.e. $A_i* \cap A_i = \varnothing$ thus $\{CS_i\}$ might be empty (this is so for noncuspidal manipulators).

Note that the characteristic surfaces are slightly different from the *pseudo-singular surfaces* defined by Tsai[5] and Miko[35] as $f^{-1}(f(A_i*))$. The pseudo-singular surfaces were defined for 3R manipulators with unlimited joints only. Moreover, they are not associated with an aspect. Thus, for a manipulator with more than two aspects, the pseudo-singular surfaces generate "spurious" surfaces that do not separate the inverse kinematics solutions. In fact, the pseudo-singular surfaces and the characteristic surfaces are equivalent for manipulators with unlimited joints and having only two aspects. Fig. 5 shows the two characteristic surfaces $\{CS_1\}$ and $\{CS_2\}$ for the illustrative manipulator.

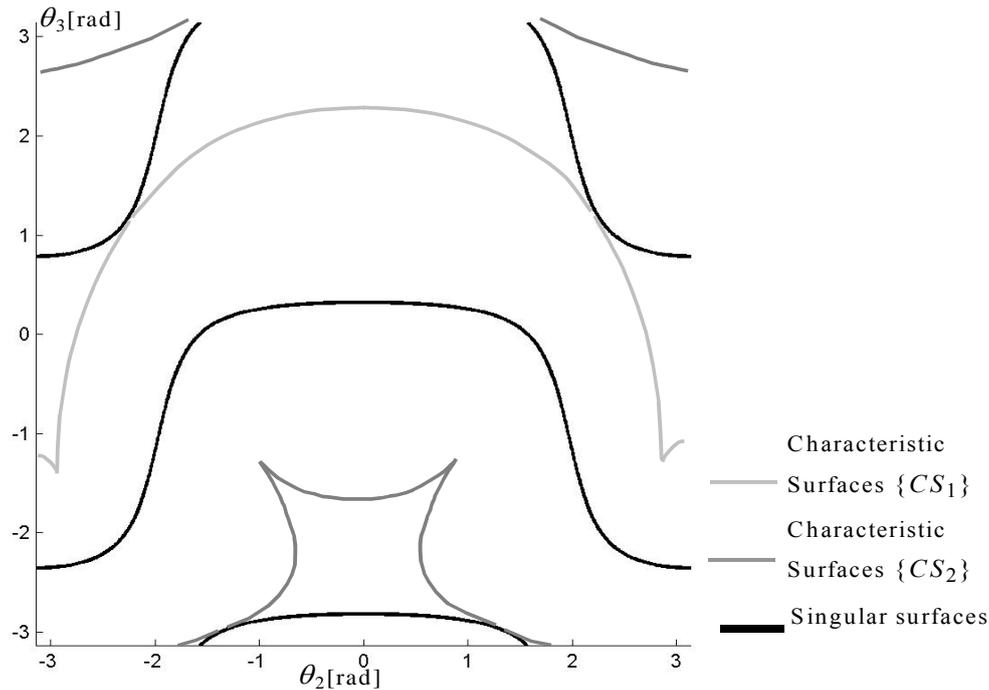

**Fig. 5. Singular and characteristic surfaces.**

The characteristic surfaces are independent of $\theta_1$ when $\theta_1$ is unlimited. Because the general definition (2) is not algebraic in nature, it is difficult to derive an algebraic expression of $\{CS_i\}$ that would be easy to handle. A scanning process can be used to plot the characteristic surfaces[19].

The characteristic surfaces induce a partition of each aspect into open sets that we call *reduced aspects*. Also, the internal boundaries induce a partition of the workspace into regions and each such region is associated with several reduced aspects. For the illustrative manipulator of Fig. 1, the inner region is associated with the four reduced aspects $Ra_{11}$, $Ra_{12}$, $Ra_{21}$ and $Ra_{22}$ (in gray in Fig. 6).

The first two, $Ra_{11}$ and $Ra_{12}$, are in $A_1$, whereas $Ra_{21}$ and $Ra_{22}$ are in $A_2$. The outer region is associated with the two reduced aspects $Ra_{13}$ and $Ra_{23}$ that belong to $A_1$ and $A_2$, respectively.

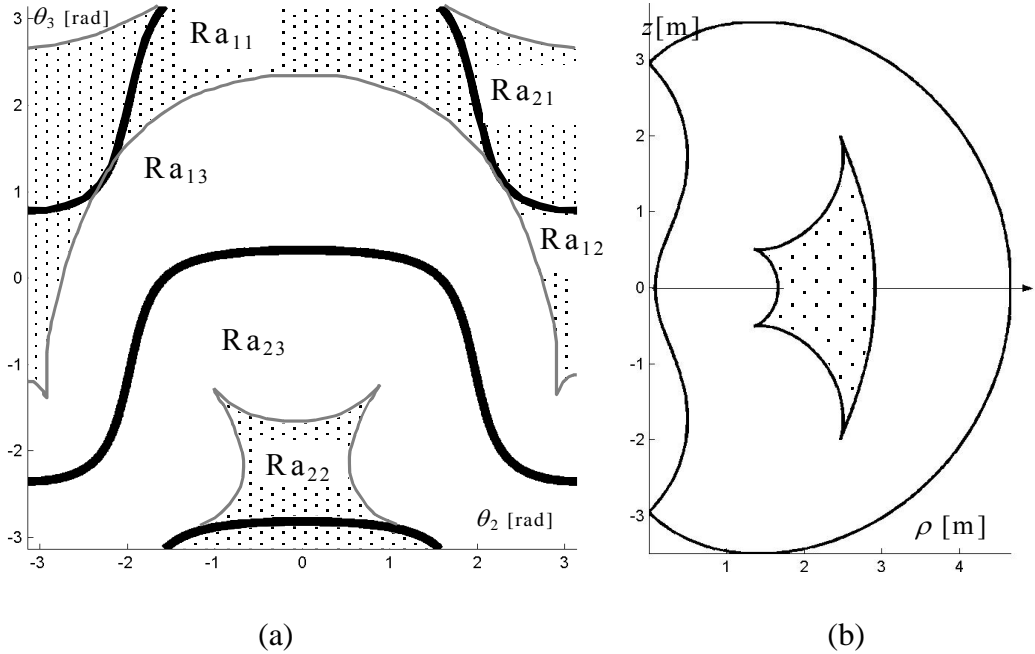

**Fig. 6. Correspondence between the reduced aspects (a) and the regions (b).**

### *3.2 Uniqueness domains*

Fig. 6 shows that each aspect is made of three reduced aspects, two of them being associated with the same region in the workspace ($Ra_{11}$ and $Ra_{12}$ for aspect $A_1$). If we remove one of these two reduced aspects and its boundary from the aspect, the remaining domain is a uniqueness domain. Thus, there is still a unique inverse kinematic solution in the domain defined by $Qu_1=A_1\dot{-}C(Ra_{12})$ as well as in $Qu_2=A_1\dot{-}C(Ra_{11})$ ($\dot{-}$ means the difference between sets, $C(R)$ means the closure of R). In the same way, $Qu_3=A_2\dot{-}C(Ra_{22})$ and $Qu_4=A_2\dot{-}C(Ra_{21})$ are still uniqueness domains. Fig. 7 shows the four maximal uniqueness domains for the illustrative manipulator.

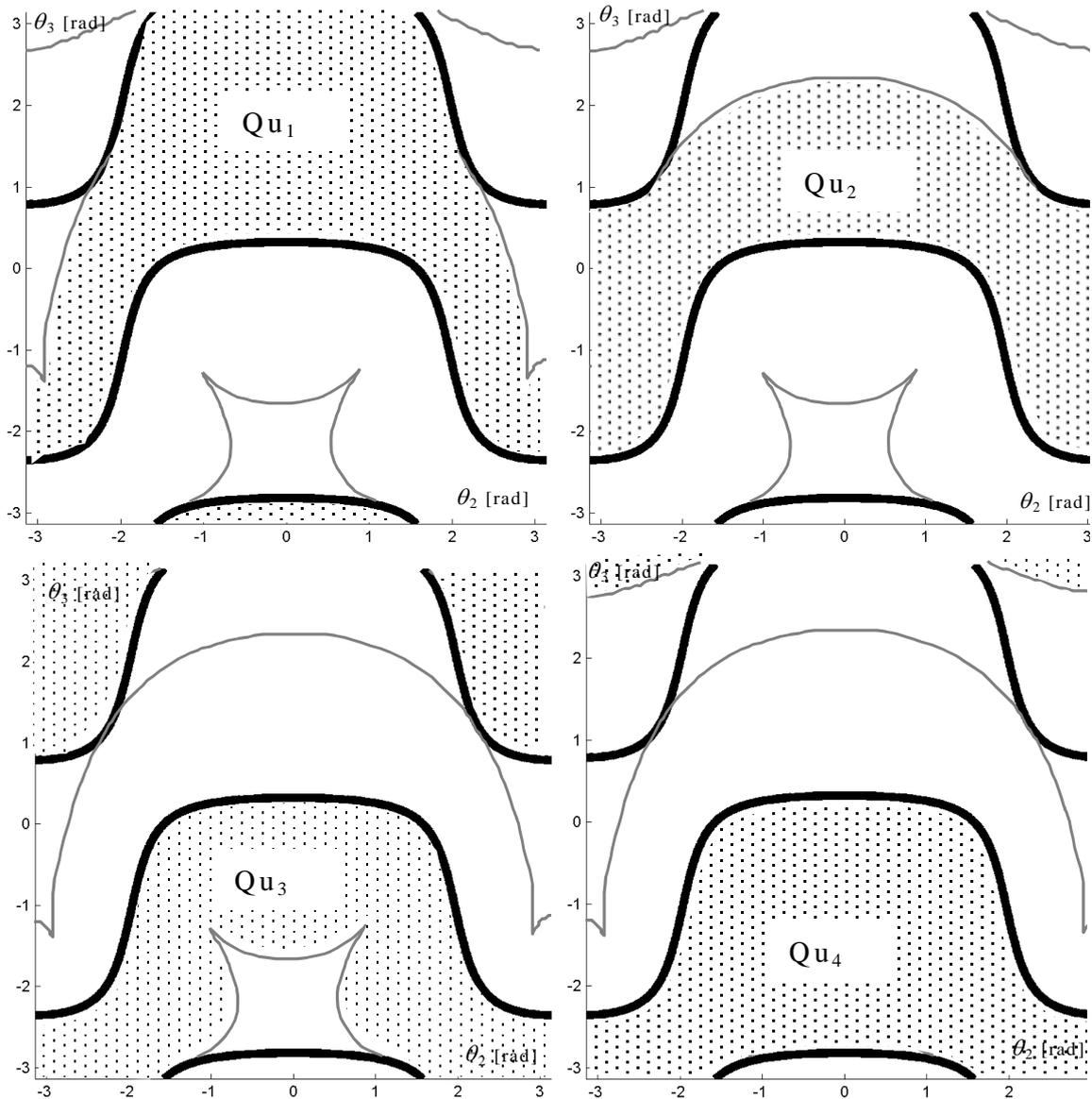

**Fig. 7. Four maximal uniqueness domains for the illustrative manipulator of Fig. 1.**

### 3.3 Regions of feasible paths in the workspace

For a noncuspidal manipulator, the regions of feasible paths in the workspace are the image of the aspects[1]. This is not true for a cuspidal manipulator because the aspects are not the uniqueness domains. The regions of feasible paths in the workspace must be defined by the uniqueness domains. Figure 8 shows the four regions of feasible paths, which are the images of the uniqueness domains in the workspace. It is important to note that $Wf_1$, $Wf_2$, $Wf_3$ and $Wf_4$ define regions where *any arbitrary path* is feasible but they do not include all feasible paths. In effect, it is possible to define a feasible path that undergoes a nonsingular change of posture in, say, aspect $A_1$ like in Fig. 4. In this case, the path will start in $Wf_1$ and stop in $Wf_2$.

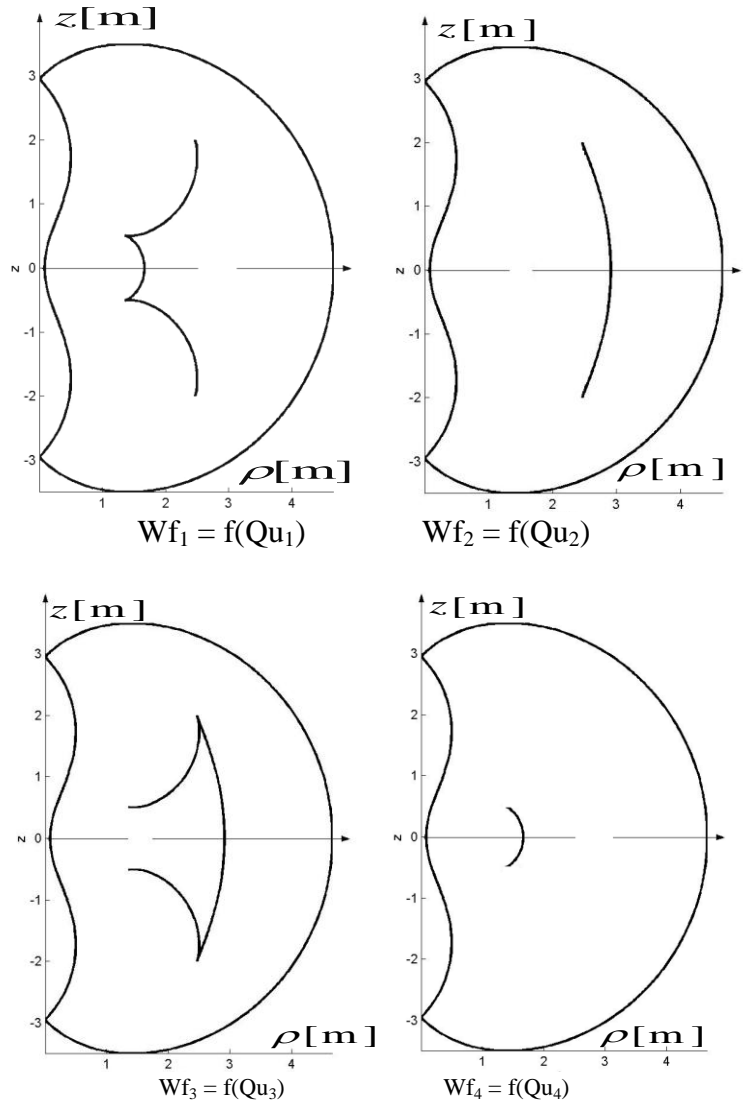

**Fig. 8. Regions of feasible paths.**

To get the full model of feasible paths in $WA_1$, $Wf_1$ and $Wf_2$ must be properly "glued" together, which can be realized by plotting the surface ($\rho$, z, cos(($\theta_2$)) for $A_1$[19]. From a mathematical point of view, this method is referred to as the level set method[37]. Also, plotting ($\rho$, z, cos(($\theta_2$)) for $A_2$ provides the full model of feasible paths in $WA_2$ (Fig. 9). This model helps better understand the nonsingular posture changing phenomenon and the loop path shown in Fig. 4.

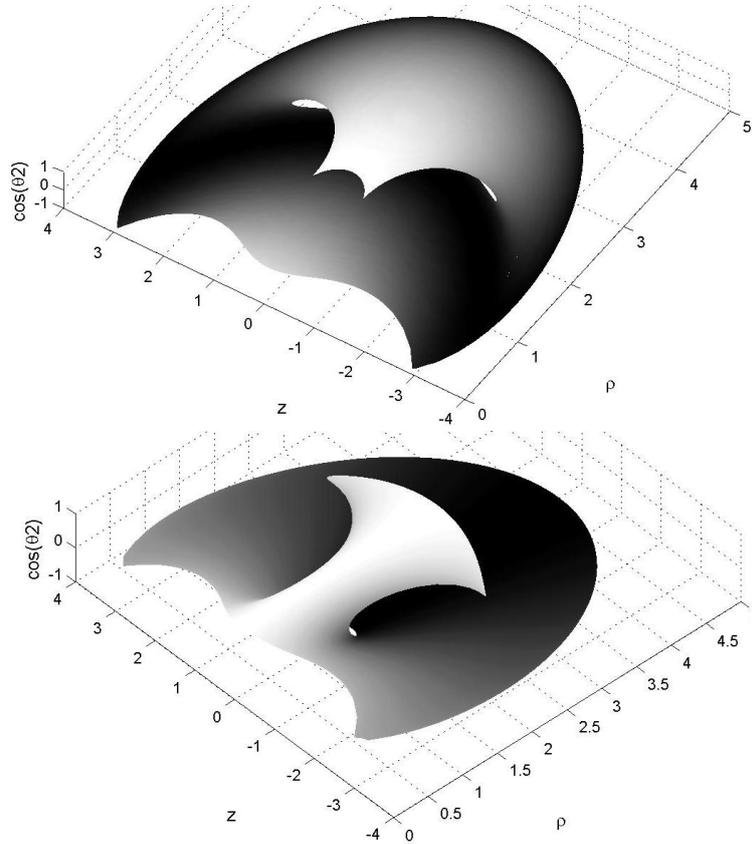

**Fig. 9. Full model for the feasible paths in $WA_1$ (above) and $WA_2$ (below).**

The uniqueness domains and the regions of feasible paths of any 3R manipulators can be got quite simply when the singular curves and the characteristic surfaces are given[19]. The regions of feasible paths are useful to assess the global performances of a manipulator and to compare several manipulator designs. These regions can also be used to verify the feasibility of a path without analyzing the root equality of the inverse kinematic polynomial on the boundaries. In effect, since each region of feasible paths is associated with one inverse kinematic solution, these regions indicate which internal surfaces can be crossed, according to the inverse kinematic solution used to follow the path[19].

The characteristic surfaces, the uniqueness domains and the regions of feasible paths can be calculated in the same way for 3R manipulators with joint limits on their last two axes[19]. When joint 1 is also limited, it is no longer possible to build 2D sections. Such manipulators were investigated by El Omri and their uniqueness domains and regions of feasible paths were built using octrees[38].

# 4 ENUMERATION OF CUSPIDAL MANIPULATORS AND CLASSIFICATION

## 4.1 Simplifying geometric conditions

Because of the more complex behavior of a cuspidal robot and because of the difficulty in modeling its kinematic properties, industrial robots should be designed noncuspidal rather than cuspidal. Thus, before designing an innovative kinematic architecture, robot manufacturers should have guidelines such as design rules to help them. Why a manipulator with a given geometry is cuspidal has long been a very intriguing question. It is worth noting that this question remains not completely solved. One of the pioneer contributors to this problem was J. Burdick, who observed that under simplifying geometric conditions such as intersecting or parallel joint axes, a 3R manipulator was noncuspidal[7]. Other simplifying conditions were exhibited later[7] as a direct consequence of the necessary and sufficient condition recalled in next section. Finally, six geometric conditions were found to define noncuspidal 3R manipulators:

    1/ first two joint axes are parallel ($\sin(\alpha_1)=0$),

    2/ last two joint axes are parallel ($\sin(\alpha_2)=0$),

    3/ first two joint axes intersect ($a_1=0$),

    4/ last two joint axes intersect ($a_2=0$),

5/ first two joint axes are orthogonal, all joint offsets are zero ($\cos(\alpha_1)=0$, $d_2=0$, $d_3=0$),

6/ joint axes are mutually orthogonal, first joint offset vanishes ($\cos(\alpha_1)=\cos(\alpha_2)=0$, $d_2=0$).

These conditions also hold for 6R manipulators with spherical wrist because the singularity analysis of the wrist can be decoupled from those of the regional structure. It is worth noting that conditions 2/ and 3/ are encountered in most industrial 6R manipulators. However, the last two conditions (5/ and 6/) are unusual.

Analogous conditions exist also for manipulators with prismatic joints[39].

### 4.2 *Necessary and sufficient condition for a manipulator to be cuspidal*

Burdick conjectured that, on the other hand, 3R manipulators with "general" geometry should be cuspidal[6]. But it was shown later that the correlation between cuspidal manipulators and general manipulators was not clear[8]. For example, the orthogonal manipulator shown in Fig. 1 is cuspidal but it is no longer cuspidal when $a_3$ is set to 0.5m instead of 1.5m (with the same values for the remaining DH-parameters). In both cases the manipulators are not of "general geometry" in the sense that the last joint offset is equal to zero and the joint axes are mutually orthogonal. An important step towards the characterization of cuspidal manipulators was established in 1995 when a general necessary and sufficient condition for a manipulator to be cuspidal was provided. This condition states that a manipulator is cuspidal if and only if its inverse kinematics admits a triple solution[9] (i.e. a point where three inverse kinematic solutions coincide). In the cross section of the workspace of a 3R manipulator, a point with three equal inverse kinematic solutions is a cusp point (see the four cusp points in Fig. 3). A direct consequence of this condition is that for a manipulator to be cuspidal, the degree of its inverse kinematics polynomial must be greater than 2. Hence, any *quadratic* manipulator (i.e. whose inverse kinematic polynomial can be reduced to a quadratics) is noncuspidal. All six noncuspidal manipulator types enumerated in the preceding section are quadratic[38]. Also, any 3-DOF manipulator (or 6-DOF with wrist) with at least two prismatic joints are always quadratic[40] and thus cannot be cuspidal. But it took eight years before this condition could be exploited to derive more general conditions on the DH-parameters[14]. The exploitation of the necessary and sufficient condition is recalled hereafter.

### 4.3 Classification of 3R orthogonal manipulators

For a 3R manipulator, the existence of cusps can be determined from its fourth-degree inverse kinematic polynomial $P(t)$ in $t = \tan(\theta_3/2)$ whose coefficients are function of the DH-parameters and of the variables $R = x^2 + y^2$ and $Z = z^2$ (see reference[30] for more details on the derivation and properties of this polynomial). The condition for $P(t)$ to have three equal roots can be set as follows:

$$\begin{cases} P(t, a_2, a_3, \alpha_1, \alpha_2, d_2, R, Z) = 0 \\ \dfrac{\partial P}{\partial t}(t, a_2, a_3, \alpha_1, \alpha_2, d_2, R, Z) = 0 \\ \dfrac{\partial^2 P}{\partial t^2}(t, a_2, a_3, \alpha_1, \alpha_2, d_2, R, Z) = 0 \end{cases} \qquad (3)$$

The three variables $t$, $R$ and $Z$ must be eliminated to obtain a condition on the DH-parameters. Deriving a symbolic solution of (3) in the general case is not reasonable and has still not been attempted. On the other hand, the study of the particular case $|\alpha_1|=\pi/2$, $|\alpha_2|=\pi/2$ (orthogonal manipulators) is interesting because this case is more tractable (although still very complex) and the family of orthogonal manipulators is rich enough to define alternative designs with relatively simple geometries that could find potential applications in industry.

Rather than simply distinguishing between cuspidal and noncuspidal orthogonal manipulators, it is more interesting to classify the family of orthogonal manipulators as function of their number of cusp points. Indeed, the number of cusps provides more information about the topology of the singular curves in the workspace and, as a result, about the global properties of the manipulator such as the existence of voids and of 4-solution regions[6,17,20,30]. To do this classification, it is appropriate to search for the conditions under which the number of real solutions of system (3) changes[41]. By doing so, a set of *bifurcating surfaces* is defined in the parameter space of orthogonal manipulators where the number of cusps changes. These bifurcating surfaces divide the parameter space into domains where all manipulators have the same number of cusp points. The bifurcating surfaces can be regarded as sets of transition manipulators. The algebra involved in system (3) is too complex to be handled by commercial computer algebra tools. Corvez and Rouillier[14] resorted to sophisticated computer algebra tools to solve system (3) by first considering the more particular case $d_3=0$ (no offset along the last joint axis like the robot shown in fig. 1). They used Groebner Bases and Cylindrical Algebraic Decomposition[41,42] to find the equations of the bifurcating surfaces and the number of domains generated by these surfaces. They obtained 5 distinct surface equations, which were shown to define 105 domains in the parameter space. A kinematic interpretation of this theoretical work was conducted by Baili et al[15] : the authors analyzed global kinematic properties of one representative manipulator in each domain. Only 5 different cases were found to exist. In fact, several surface equations obtained by Corvez and Rouillier were extraneous solutions. Finally, the true bifurcating surfaces were shown to take on the following explicit form[15]:

$$C_1: a_3 = \sqrt{\frac{1}{2}(a_2^2 + d_2^2) - \frac{(a_2^2 + d_2^2)^2 - (a_2^2 - d_2^2)^2}{AB}} \quad (4)$$

$$C_2: a_3 = \frac{a_2}{1+a_2} A \quad (5)$$

$$C_3: a_3 = \frac{a_2}{a_2 - 1} B \text{ and } a_2 > 1 \quad (6)$$

$$C_4: a_3 = \frac{a_2}{1 - a_2} B \text{ and } a_2 < 1 \quad (7)$$

where $\quad A = \sqrt{(a_2 + 1)^2 + d_2^2} \text{ and } B = \sqrt{(a_2 - 1)^2 + d_2^2} \quad (8)$

Note that the above equations assume $d_3=0$. Moreover, $a_1$ was set to 1 without loss of generality in order to handle only three independent parameters. These four surfaces divide the parameter space into 5 domains with 0, 2 or 4 cusps. Fig. 10 shows the plots of the surfaces in a section ($a_2$, $a_3$) of the parameter space for $d_2=1$. Plotting sections for different values of $d_2$ changes the size of each region but the general pattern does not change and the number of cells remains the same. There are two domains of noncuspidal manipulators (domains 1 and 5), two domains of manipulators with four cusps (domains 2 and 4) and one domain of manipulators with two cusps.

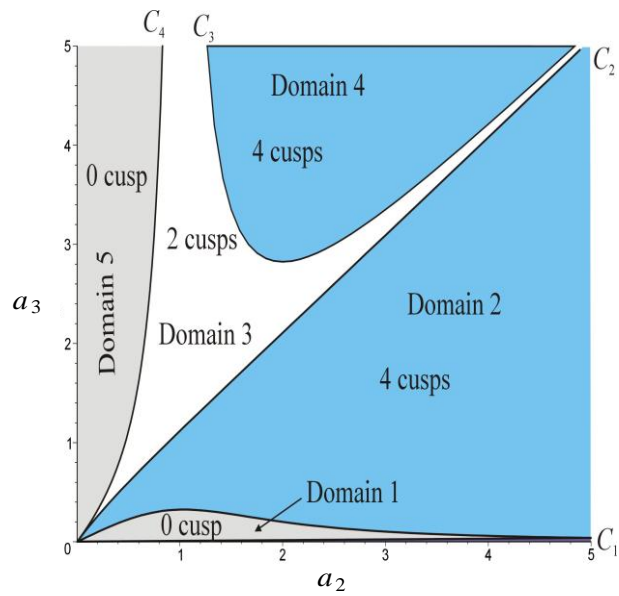

**Fig. 10. Plots of the bifurcating surfaces in a section ($a_2$, $a_3$) of the parameter space for $d_2=1$.**

The partition of the parameter space and the equations of the bifurcating surfaces allow us to define an explicit necessary and sufficient condition for an orthogonal manipulator with no offset along its last joint axis to be cuspidal. In effect, figure 10 shows that a manipulator is cuspidal if and only if it belongs to domains 2, 3 or 4. Thus, an orthogonal manipulator with no offset along its last joint axis is cuspidal if and only if:

$$a_3 \geq \sqrt{\frac{1}{2}(a_2^2 + d_2^2 - \frac{(a_2^2 + d_2^2)^2 - a_1^2(a_2^2 - d_2^2)}{\sqrt{(a_2+a_1)^2 + d_2^2}\sqrt{(a_2-a_1)^2 + d_2^2}})} \quad (9)$$

and

$$a_2 > a_1 \text{ or } (a_2 < a_1 \text{ and } a_3 = \frac{a_2}{a_1 - a_2}\sqrt{(a_2 - a_1)^2 + d_2^2}) \quad (10)$$

Note that parameter $a_1$ was no longer set equal to 1 in order to show better the influence of all parameters. Fig. 11 shows the cross sections of the workspace and the singular curves in the joint space, for one representative manipulator in each domain of the partition. The number of inverse kinematic solutions in each region of the workspace is indicated. Fig. 11 shows that manipulators in domain 1 have only two inverse kinematics solutions. Also, they have a void in their workspace and they are noncuspidal. In fact, it can be shown that all other manipulators have 4 inverse kinematic solutions and that Eq. (9) is a necessary and sufficient condition for an orthogonal manipulator with $d_3=0$ to have four inverse kinematic solutions[20]. The other noncuspidal manipulators are in domain 5. They have a region with 4 inverse kinematic solutions and no void.

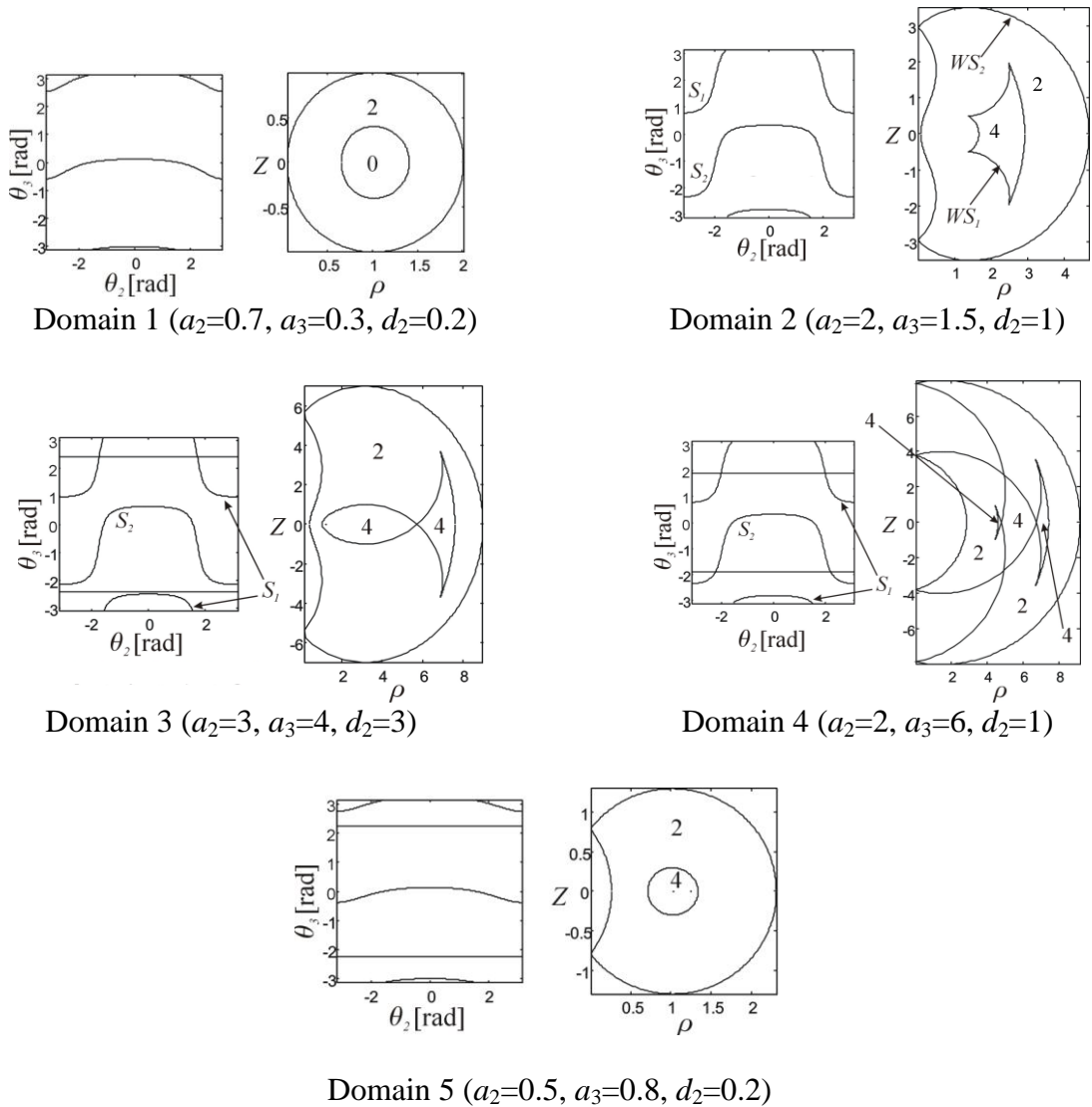

Domain 1 ($a_2=0.7$, $a_3=0.3$, $d_2=0.2$)    Domain 2 ($a_2=2$, $a_3=1.5$, $d_2=1$)

Domain 3 ($a_2=3$, $a_3=4$, $d_2=3$)    Domain 4 ($a_2=2$, $a_3=6$, $d_2=1$)

Domain 5 ($a_2=0.5$, $a_3=0.8$, $d_2=0.2$)

**Fig. 11. Workspace topologies in each domain.**

Each domain of the parameter space can be further classified by taking into account the number of nodes, i.e., the number of intersection points on the workspace boundaries[17]. There is one node in the workspace of the illustrative manipulator of domain 3 (fig. 11) but two more nodes may arise when the internal boundary goes outside the external one (as displayed in Fig. 13, $WT_6$). Two more surface equations $E_1=\frac{1}{2}(A-B)$, $E_2=a_2$ and $E_3=\frac{1}{2}(A+B)$ ($A$ and $B$ are defined as in (8)) appear when the number of nodes is considered and the parameter space is divided into 9 cells, each one being associated with a particular workspace topology $WT_i$. This new partition is shown in a section $(a_2, a_3)$ of the parameter space for $d_2=1$ (Fig. 12).

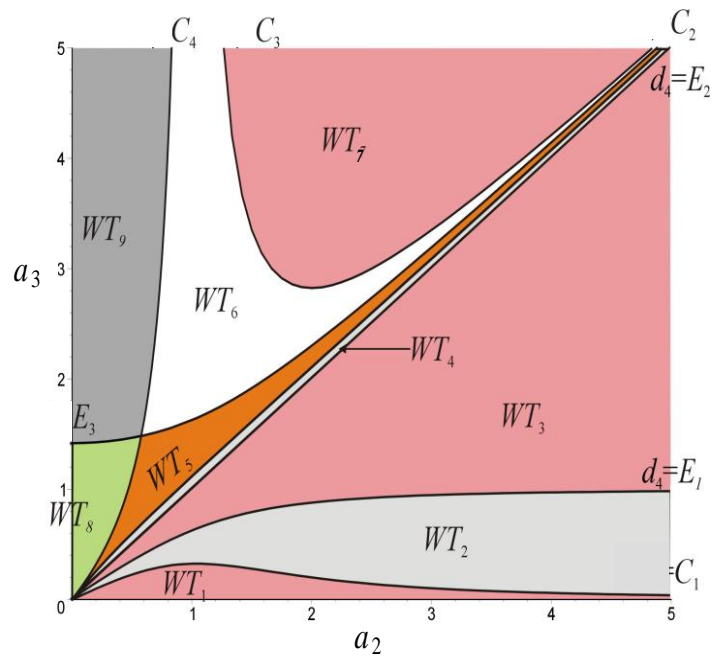

$WT_1$: 0 cusp, 0 node, $WT_2$: 4 cusps, 2 nodes, $WT_3$: 4 cusps, 0 node

$WT_4$: : 4 cusps, 2 nodes, $WT_5$: 2 cusps, 1 node, $WT_6$: 2 cusps, 3 nodes,

$WT_7$: 4 cusps, 4 nodes, $WT_8$: 0 cusp, 0 node, $WT_9$: 0 cusp, 2 nodes

**Fig. 12. Partition of the parameter space according to the number of cusps and nodes**.

Plots of the separating curves in sections for different values of $d_2$ show that they deform smoothly with the same intersections when $d_2$ varies. The areas of $WT_1$, $WT_2$, $WT_7$ and $WT_9$ increase when $r_2$ decreases, whereas those of $WT_3$, $WT_4$, $WT_5$ and $WT_6$ decrease. All manipulators in domain 1 have no node (this domain is referred to as $WT_1$), all manipulators in domains 4 have four nodes ($WT_7$). Manipulators in domains 2 and 5 may have no nodes ($WT_3$ and $WT_8$, respectively) or two nodes ($WT_2$ and $WT_9$, respectively) and manipulators in domain 3 may have one or three nodes ($WT_5$ and $WT_6$, respectively).

The classification with nodes makes it possible to identify all the orthogonal manipulators that have a void in their workspace. In effect, all such manipulators are in domains $WT_1$ and $WT_2$.

Fig. 13 shows the cross sections of the workspace and the singular curves in the joint space, for one representative manipulator in $WT_2$, $WT_4$, $WT_6$ and $WT_9$. The other representative manipulators appear in fig. 11. The two horizontal singular lines that appear in the joint space of the manipulators in $WT_4$ and $WT_6$ generate isolated points in the workspace cross section as shown by arrows in Fig. 12.

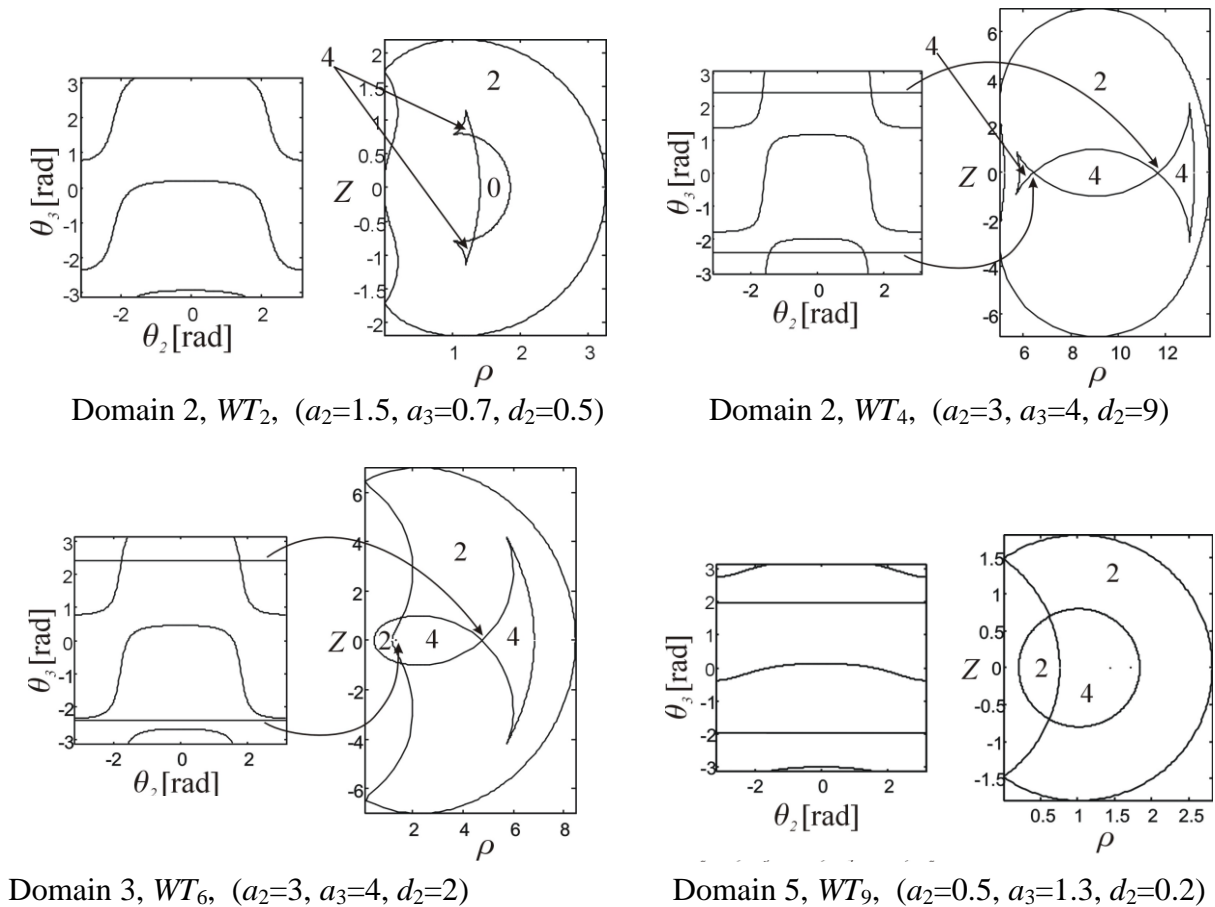

Domain 2, $WT_2$,  ($a_2$=1.5, $a_3$=0.7, $d_2$=0.5)          Domain 2, $WT_4$,  ($a_2$=3, $a_3$=4, $d_2$=9)

Domain 3, $WT_6$,  ($a_2$=3, $a_3$=4, $d_2$=2)          Domain 5, $WT_9$,  ($a_2$=0.5, $a_3$=1.3, $d_2$=0.2)

**Fig. 13. Workspace topologies $WT_2$, $WT_4$, $WT_6$ and $WT_9$.**

The preceding classification stands for orthogonal manipulators with no offset along their last joint axis ($d_3=0$). For manipulators with nonzero offsets, it is possible to solve system (3) as above but the resulting separating surface equations get very complicated and the parameter space is four-dimensional[18]. The complete classification for general 3R orthogonal manipulators was conducted in Baili's PhD thesis[43]. The classification shows that manipulators may have 0, 2, 4, 6 or 8 cusps. The equation of one of the bifurcating surfaces between domains with different number of cusps is a 12$^{th}$-degree polynomial in the square of the DH-parameters and contains 536 monomials. It is interesting to remark that when $d_3$ is set to zero, this equation simplifies considerably and factors. Equating the three factors to zero gives exactly the three equations (5), (6) and (7).

For small values of $d_3$, the partition sections look like those in Fig. 12 but the subspace $WT_4$ does not exist any more. It is replaced by two adjacent subspaces with 6 and 8 cusps, located near $a_3=a_2$. For high values of $d_3$, the partition gets very complicated with not less than 22 distinct topologies. Figure 14 shows an example of a workspace topology with 8 cusps and 4 nodes. The zoomed view shows the creation of a pair of cusps, a pattern known as a swallowtail[44].

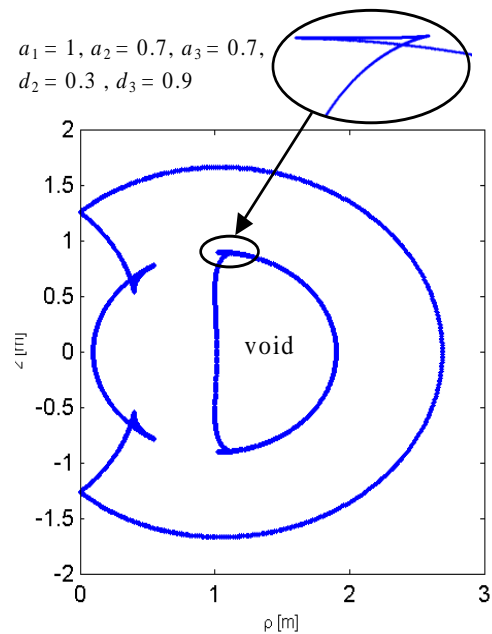

**Fig. 14. Workspace topology when $d_3 \neq 0$.**

*Note1*: all results about the classification presented in this section assume that $a_1$, $a_2$, $a_3$ and $d_2$ are different from zero. In any other case, the manipulator is noncuspidal (this is because if one of these parameters is equal to zero, one of the simplifying conditions given in section 3.1 is satisfied). However the classification according to the number of nodes is still possible and was conducted recently by Zein et al[45]. The classification revealed interesting noncuspidal 3R orthogonal architectures with good workspace properties (workspace of regular shape, fully reachable with four inverse kinematic solutions and in which any path is feasible) that could find industrial applications.

*Note2*: the classification of 3-DOF manipulators with one prismatic joint can be attempted with the same tools as for 3R manipulators. Because the kinematic equations are simpler when a prismatic joint is involved, the classification would be simpler.

## 5   Some facts about cuspidal 6R robots

The above classifications also hold for 6R manipulators with spherical wrist (i.e, with their last three joint axes intersecting at a common point) because the singularity analysis of the wrist can then be decoupled from that of the regional structure. In the introduction of this paper, we reported the story of the IRB 6400C robot. This robot, shown in Fig. 15, has a spherical wrist and its regional structure is an orthogonal 3R manipulator that can be shown to be cuspidal[43]. Note that the main objective of this new robot design was to save space and this is why its first joint axis is horizontal instead of vertical[46]. This was a good idea but at the time the engineers of ABB designed their new robot, the classification results were not published. It would be interesting to attempt a new design, keeping the orthogonal architecture with its first axis horizontal but tuning the length parameters in order that the robot falls in one of the interesting classes of noncuspidal orthogonal manipulators described by Zein et al[45].

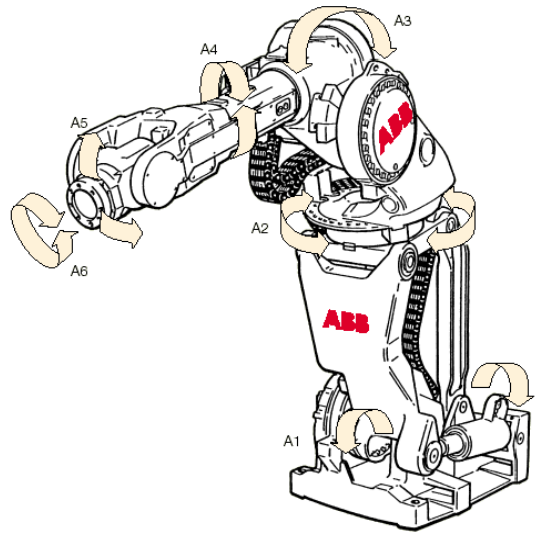

**Fig. 15. The ABB IRB 6400C robot.**

On the other hand, there is no general result about the enumeration of cuspidal 6–DOF manipulators with nonspherical wrist. One of the reasons is the difficulty in analyzing the singularities of general 6R robots, which depend on four joint variables instead of two in 3R robots. We think that 6R manipulators are very likely to be cuspidal, even if the simplifying geometric conditions listed in section 4.1 are satisfied. This is because the inverse kinematics of most 6-DOF manipulators with nonspherical wrist is a polynomial of degree higher than 4, which may admit triple roots. Further research work is required before stating more definitive results but several examples of simple 6R robots with nonspherical wrist exist. One of these robots is the GMF P150 shown in Fig. 16 used in the automotive industry for car painting (a similar version exists by COMAU). This robot is close to a PUMA robot, the only difference being the presence of a wrist offset. El Omri showed that without taking account the joint limits, this robot has 16 inverse kinematic solutions and only two aspects[38]. Thus, it is cuspidal. Another example is the ROBOX painting robot studied by M. Zoppi[47] (Fig. 17). The kinematic architecture is very close to the GFM P150 but the wrist offset is not along the same wrist axis. This cuspidal robot has also 16 inverse kinematic solutions and only two aspects.

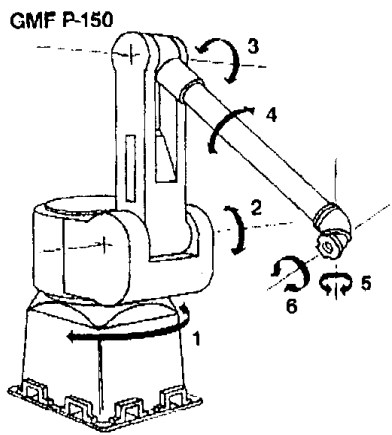 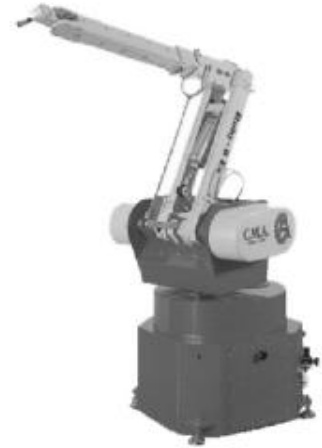

**Fig. 16. The GMF P150 robot.**        **Fig. 17. The ROBOX robot.**

If the enumeration of 6-dof cuspidal and noncuspidal manipulators is far from being established, it is possible to enumerate a set of noncuspidal ones, namely, those whose inverse kinematics polynomial is a quadratics (because no cusp point exists in this case). Such manipulators were enumerated by Mavroidis and Roth in 1996[48].

## 6    Concluding remarks

This synthesis article on cuspidal manipulators can be summarized as follows.

Cuspidal manipulators, which were first discovered in 1988, have multiple inverse kinematic solutions (IKS) that are not separated by a singular surface. In the joint space, additional surfaces, called the characteristic surfaces, divide the aspects and separate the IKS. These surfaces are used to define new uniqueness domains and regions of feasible paths in the workspace. The definitions are general and stand for any serial, nonredundant manipulator with or without joint limits. For 3-DOF manipulators, it is possible to calculate and plot these characteristic surfaces, uniqueness domains and regions of feasible paths. If the first joint is revolute and unlimited, 2-dimensional plots are sufficient. Because there is no simple algebraic definition, these sets must be calculated numerically.

A 3R manipulator is noncuspidal as soon as its first two or last two joint axes are parallel or intersect, or if the three joint axes are mutually orthogonal and the *first* joint offset is equal to zero. But an orthogonal manipulator with its *last* joint offset equal to zero may be cuspidal.

A manipulator is cuspidal if and only if there is at least one point where three inverse kinematic solutions coincide. For a 3-DOF manipulator, this point appears as a cusp point in a cross section of the workspace. This necessary and sufficient condition for a manipulator to be cuspidal makes it possible to classify orthogonal manipulators as function of the number of cusps and nodes in the workspace. For orthogonal manipulators with no joint offset along their last axis, this classification enables one to derive explicit DH-parameter based necessary and sufficient conditions for a manipulator to be cuspidal, to have four inverse kinematic solutions or to have a void in its workspace. For general 3R orthogonal manipulators, the classification is much more complex and does not lend itself to explicit conditions.

Little research work has been conducted on 6R cuspidal manipulators. It appears that 6R robots with nonspherical wrist are very likely to be cuspidal, even if two joint axes intersect or are parallel. However, there is still much work to do before having definitive geometric conditions for general 6R robots. Resorting to some transversality theorems used by singularity theorists would help going further, providing that we remain in the generic case[16,49]. So the first step would be to enumerate all 6R generic manipulators.

The case of parallel manipulators has not been considered in this article. As first observed in 1998, a parallel manipulator may be cuspidal in the sense that it may change its assembly-mode without crossing a parallel-type singularity[50, 51]. As shown by R. McAree[52] and explained in details by Zein[53], to be cuspidal, a parallel manipulator should have 3 coincident assembly modes, which define a cusp point in a section of its joint space. Because the kinematic equations of a parallel manipulator are very complex, it seems very difficult to derive general geometric conditions for a parallel manipulator to be cuspidal. To the author's knowledge, the only available results pertain to planar manipulators : to be cuspidal, a 3-DOF parallel planar manipulator with three R$\underline{P}$R legs should not have similar platform and base triangles[52, 54] but this is false if the legs are $\underline{R}$RR instead of R$\underline{P}$R[55] (the underlined letter refers to the actuated joint).

## Acknowledgment

Most of the classification work summarized in this article has been conducted in collaboration with IRMAR (Institute of Applied Mathematics of Rennes), INRIA (Research Institute of Informatics and Automatics), and LIP6 (Informatics Laboratory of Paris VI), in the frame of the national program MATHSTIC entitled "Cuspidal robots and triple roots" funded by C.N.R.S. (French Council of Scientific Research) and French Ministry of Education and Research.